%% file: main.tex
\documentclass[conference]{IEEEtran}
\IEEEoverridecommandlockouts
\usepackage{cite}
\usepackage{amsmath,amssymb,amsfonts}
\usepackage{tabularx}
\usepackage{booktabs}
\usepackage{enumitem}
\usepackage{subcaption}
\usepackage{graphicx}
\usepackage{textcomp}
\usepackage{graphicx}
\usepackage{epsfig}
\usepackage{textcomp}
\usepackage{xcolor}
\usepackage{mathrsfs}
\usepackage{eucal}
\usepackage{pifont}
\usepackage{amsthm}
\usepackage{float}
\usepackage{color}
\usepackage{dsfont}
\usepackage{algorithm}
\usepackage{algpseudocode}

\def\BibTeX{{\rm B\kern-.05em{\sc i\kern-.025em b}\kern-.08em
    T\kern-.1667em\lower.7ex\hbox{E}\kern-.125emX}}
\begin{document}
\title{\vspace{7.8mm} \LARGE \bf Smooth Computation without Input Delay: Robust Tube-Based Model Predictive Control for Robot Manipulator Planning \\
\thanks{Y. Luo, Q. Sima, T. Ji, H. Liu and F. Sun are with the Department of Computer Science and Technology, Tsinghua University, Beijing, China. J. Zhang is with Department of Informatics, University of Hamburger,Hamburg, Germany. $^{*}$Both authors contributed equally to this research, $^{\dagger}$Corresponding author Fuchun Sun (\texttt{fcsun@tsinghua.edu.cn})}
}

\author{\IEEEauthorblockN{Yu Luo$^{*}$, Qie Sima$^{*}$, Tianying Ji, Fuchun Sun$^{\dagger}$, Huaping Liu, Jianwei Zhang}
}

\maketitle

\begin{abstract}
Model Predictive Control (MPC) has exhibited remarkable capabilities in optimizing objectives and meeting constraints. However, the substantial computational burden associated with solving the Optimal Control Problem (OCP) at each triggering instant introduces significant delays between state sampling and control application. These delays limit the practicality of MPC in resource-constrained systems when engaging in complex tasks. The intuition to address this issue in this paper is that by predicting the successor state, the controller can solve the OCP one time step ahead of time thus avoiding the delay of the next action. To this end, we compute deviations between real and nominal system states, predicting forthcoming real states as initial conditions for the imminent OCP solution. Anticipatory computation stores optimal control based on current nominal states, thus mitigating the delay effects. Additionally, we establish an upper bound for linearization error, effectively linearizing the nonlinear system, reducing OCP complexity, and enhancing response speed. We provide empirical validation through two numerical simulations and corresponding real-world robot tasks, demonstrating significant performance improvements and augmented response speed (up to $90\%$) resulting from the seamless integration of our proposed approach compared to conventional time-triggered MPC strategies.
\end{abstract}


\input{01intro}

\input{02related}
\input{03problem}
\input{04method}
\input{05experiment}

\begin{figure}[t]
    \centering
    \includegraphics[width=1.0\linewidth]{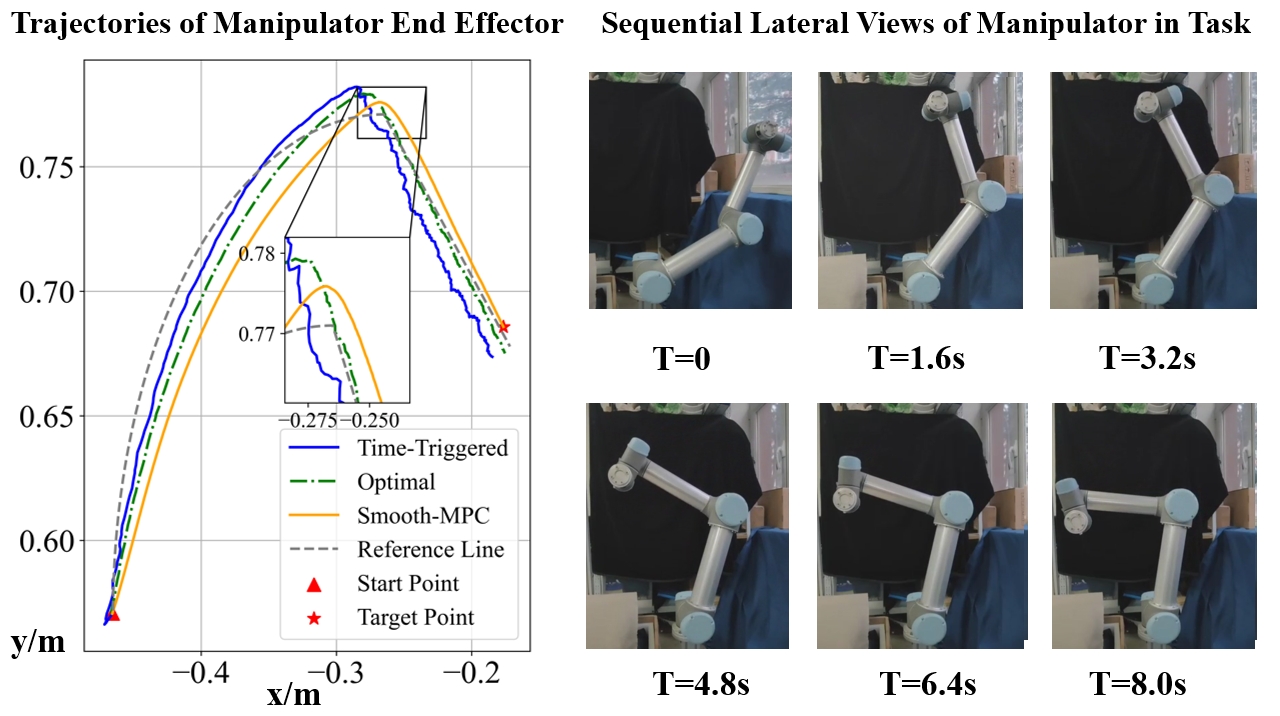}
    \caption{The trajectories of the end-effector in trajectory tracking task with different control strategies and video frames of experimental scenario.}
    \label{fig:real-trajectory}
    \vspace{-10pt}
\end{figure}

\section{Conclusion}
\label{sec:future work}
In this paper, we aim to eliminate the delay caused by solving the OCP in disturbed nonlinear systems and propose a novel robust tube-based smooth-MPC to ensure optimal control performance. We estimated the linearization error as a bounded disturbance to linearize the nonlinear system and reduce the computational complexity of the OCP. Then, the deviation of the nominal system and the real system states are deduced by Lipschitz continuity and triangle inequations to predict the region of the next real states. Based on these two mechanisms, the difficulties of the delay by solving OCP in fast dynamic systems are dramatically disposed and the optimality of this controller is guaranteed. We compare Smooth-MPC with baselines through two fundamental manipulation tasks both in simulation and real-world scenarios. Both simulation and real-world experimental results verify the control performance, fast response speed and robustness of our proposed method. Our future work will concentrate on the dynamic system of robot manipulators with stochastic disturbances to extend the application of this control method.

\section*{Acknowledgments}
This work was jointly funded by the National Science and Technology Major Project of the Ministry of Science and Technology of China (No.2018AAA0102900) and ``New Generation Artificial Intelligence'' Key Field Research and Development Plan of Guangdong Province (No.2021B0101410002).

\clearpage


\bibliographystyle{IEEEtran}
\bibliography{refs}

\end{document}

%% file: 01intro.tex
\section{Introduction}
Robot manipulator planning is a burgeoning field at the intersection of robotics, artificial intelligence, and engineering which refers to the agile and precise handling of objects by robotic systems\cite{lozano1987simple,zhu2019robotic,billard2019trends,guo2020repetitive}. In the realm of manipulation planning, control strategies play a pivotal role in enabling robots to perform intricate tasks with precision and adaptability. These strategies encompass a diverse range of methodologies which can be categorized into 3 main kinds: PID-based control, optimization-based control\cite{wang2021efficient,ding2021representation} and learning-based control\cite{omer2021model,hewing2020learning}. Model Predictive Control (MPC), a predictive control strategy\cite{MAYNE20142967,2020Industry}, holds a significant place in optimization-based control methods, by optimizing system targets and simultaneously managing state and control input constraints\cite{2016Model, BrunnerBKPZNS20, HanniganSKHYC20}.

When applying MPC, it may present several disadvantages including excessive computational complexity, high computational delay\cite{kim2019highly} and sensitivity to parameters.  Particularly, computational delay, often encountered in complex robotic systems, can render significant impacts on the performance of MPC in dexterous manipulation tasks. High computational delay can lead to error accumulation over time, causing the robot to deviate from its intended trajectory or even constraint violations\cite{schwenzer2021MPC}. Addressing these issues in the context of fast dynamic systems and resource-limited platforms has spurred impressive efforts, primarily manifesting in two approaches: reducing computational complexity\cite{Han2020local,GRIFFITH2018109,lizhao9007513} or diminishing OCP solving frequency\cite{LiS14a,liu8010327,zhao2019event}. However, these efforts address computational complexity, the single solving time at each update instance often falls short of real-time requirements.

\begin{figure}[t]
	\centering
	\includegraphics[width=1.0\linewidth]{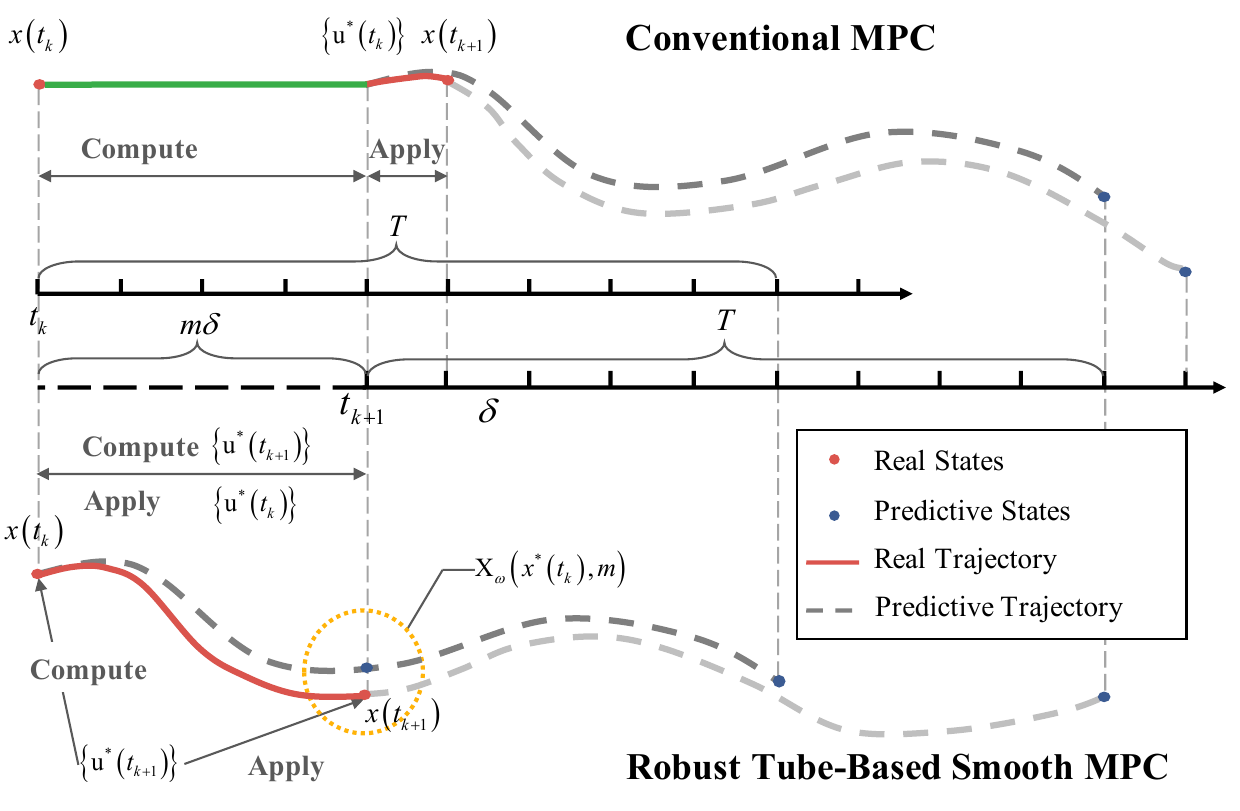}
	\caption{In this figure, we compare the implementation of the conventional MPC with our designed approach. Compared with the conventional control fashion, our approach predicts the next real states $x(t_{k+1})$ by the estimated predictive disturbed state set $\mathbb{X}_{\omega}(x^*(t_k),m)$ to obtain the next optimal control input $\textbf{u}^*(t_{k+1})$. Thus, when the next real states $x(t_{k+1})$ come, we can directly use $\textbf{u}^*(t_{k+1})$ and compute $\textbf{u}^*(t_{k+2})$ to achieve smoothness. Compared with the conventional MPC, our approach improves the response speed and keeps the optimal control performance.}
	\label{fig:mcmthesis-logo}
    \vspace{-15pt}
\end{figure}

In this paper, we modify the pace of OCP solving and control inputs to mitigate the delay control input, and propose a novel framework, robust tube-based smooth MPC, for robotics manipulator planning with constraints and disturbances, as illustrated in Fig. \ref{fig:mcmthesis-logo}. Specifically, in a departure from prior works, our method foresees the next real state region by solving the OCP ahead at the present moment, grounded in nominal predictions. Leveraging nonlinear system linearization, we mitigate computational intricacies of a single time of the OCP solution, and harness the robustness framework inherent to tube MPC. The crux of our contributions encompasses three key aspects:
\begin{itemize}
    \item We establish a predictive disturbed state set for forthcoming real states, serving as the foundational initial condition for the succeeding OCP at the present instant. This anticipatory stride effectively circumvents computational delays, improving the smoothness of the MPC framework.
    \item To enhance solution speed by diminishing the nonlinear OCP solving complexity, we employ piece linearization technology. This transformative approach translates the nonlinear OCP into a linear counterpart, thereby facilitating efficient solution determination.
    \item We corroborate our approach's efficacy through both simulation and real-world system validations. Comparative analyses against ideal MPC and time-triggered MPC methodologies reveal heightened response speed and superior optimal performance.
\end{itemize}

%% file: 02related.tex
\section{Related Work}
\label{sec:related work}
\subsection{Model Predictive Control}
As discussed in the previous section, previous works have focused on improving the response speed of MPC in two avenues: reduced complexity or decreased frequency. For the first avenue, Han and Tedrake \cite{Han2020local} utilize piecewise linear affine approximation in dexterous robotic manipulation, particularly when facing non-smooth nonlinear systems and significant external disturbances. Another effective technique, aside from model linearization, involves curtailing prediction horizons to trim OCP computation time \cite{GRIFFITH2018109,lizhao9007513}. On the other hand, lowering OCP solving frequency has gained attention, leading to the development of event-triggered MPC and self-triggered MPC. Li and Liu \cite{LiS14a,liu8010327} explore event-triggered MPC's continuous-time nonlinear systems application, a concept extended to modular reconfigurable robots' decentralized tracking control in \cite{zhao2019event}. 

Although these efforts address computational complexity, the computation time for solving the OCP in a single step remains excessively protracted to meet the transient control period requisites in swiftly evolving systems\cite{fei2021real, ding2021representation, kleff2021high, wang2021efficient}. Consequently, this temporal lag between sampling and input provisioning induces delays, compounding the issue. Even after the control input is computed, real system states have often changed, creating a discordance between state and input. In light of this practical quandary, efforts in this paper have emerged to eliminate delay through asynchronous sampling and input, leveraging MPC's predictive capabilities.


\subsection{Manipulator Motion Planning}

The robotics community has studied on motion planning algorithms for manipulators from very early times\cite{lozano1987simple,faverjon1987local,lengyel1990real}. Up to the present, various strategies spanning from early path planning methods to recent advancements in adaptive and data-driven techniques have been proposed\cite{zhou2022review}. The early works in manipulator motion planning primarily focused on finding a collision-free trajectory between the initial and the target. Methods such as the rapidly-exploring random tree (RRT)\cite{rodriguez2006obstacle} and probabilistic roadmaps (PRM)\cite{missiuro2006adapting,simeon2000visibility} solve this problem by providing efficient solutions to high-dimensional configuration spaces. 
Then, researchers switch to trajectory optimization and model predictive control (MPC) to fulfil the requirements of real-time planning in dynamic environments. Optimization-based approaches have been proposed to enable manipulators to adapt to unforeseen obstacles and constraints while maintaining a high response speed\cite{ding2021representation,wang2021efficient}. In recent years, learning-based motion planning methods have surged.  Neural network-based planners\cite{qureshi2020motion,xu2021motion,silver2021planning} and reinforcement learning\cite{cai2021modular}, have demonstrated remarkable capabilities in learning complex motion policies and optimizing control strategies.\par

Despite the significant progress mentioned above, several challenges persist in this domain, including real-time planning, handling uncertainty scenarios, and ensuring robustness under unforeseen disturbances. Furthermore, issues related to safe human-robot interaction also demand attention\cite{ichnowski2020deep}.

%% file: 03problem.tex
\section{Problem Formulation}
\label{sec:problem}
In this paper, we consider a discrete-time perturbed nonlinear system with state and input constraints as
\begin{eqnarray}\label{sys}
x(t+1) = f(x(t),u(t)) + \omega(t), \  t \in \mathbb{N}_{\geqslant 0},
\end{eqnarray}
where $x(t) \in \mathbb{R}^n$ and $u(t) \in \mathbb{R}^m$ represents the system states and the control inputs. For this nonlinear system, $\omega(t) \in \mathbb{W}(t)\cap\{0\}$ means the additional disturbances or the model uncertainty, which is bounded by $\|\omega_t\|\leqslant\eta_1$. In the real control process, the system is subjected to the following hard constraints as
\begin{eqnarray}
x(t) \in \mathbb{X} \subseteq \mathbb{R}^n, \quad u(t) \in \mathbb{U} \subseteq \mathbb{R}^m,
\end{eqnarray}
where the set $\mathbb{X}$ and $\mathbb{U}$ are convex and compact including the origin as an interior point. Further, the nominal system of (\ref{sys}) is introduced as
\begin{eqnarray}\label{nonminalsys}
x(t+1) = f(x(t),u(t)).
\end{eqnarray}
The following reasonable assumptions are given to describe the properties of the system model in two aspects about linearising the nonlinear systems in piece-wise control period and the Lipschitz continuity.

\subsection{System Hypothesis}
To linearize this disturbed nonlinear system piece-wise, we give two assumptions of the model $f(x,u)$ below.

\textbf{Assumption 1:} the function $f$ is a twice continuously differentiable function and $f(0,0)=0$. Therefore, the nominal system model can be linearized at each sampling instant $t_k$ as:
\begin{eqnarray}
A_{t_k} = \frac{\partial f}{\partial x}\big|_{(x(t_k),u(t_k))},\quad B_{t_k} = \frac{\partial f}{\partial u}\big|_{(x(t_k),u(t_k))}.
\end{eqnarray}
Further, considering the piece-wise control interval $[t_k,t_{k+1}]$, the Hessian matrix $H(x,u)$ is bound as
\begin{equation}
    \|H(x(t),u(t))\|  \leqslant \eta_H.
\end{equation}

Thus, the system matrix pair $(A_{t_k},B_{t_k})$ is stabilizable and a state-feedback gain $K$ can be chosen as the local controller $u(t)=Kx(t),t\in[t_k,t_{k+1}]$, which is a feasible control solution in the control period. On the other hand, for the general real system, like robot manipulator and vehicle driving, the system models are smooth functions seldom with cuspidal points for a piece control interval to explain the reasonable bound of the Hessian matrix $H(x,u)$.

\textbf{Assumption 2:} $f(x,u)$ is locally Lipschitz continuous with respect to $x$ and $u$. With the control inputs $u_1, u_2\in \mathbb{U}$, $\forall x_1, x_2 \in \mathbb{X}$ the system satisfies: $\| f(x_1,u_1)-f(x_2,u_2) \| \leqslant l_1 \|x_1-x_2\| + l_2 \|u_1-u_2\|$. 

By this assumption, the nonlinearity of the system (\ref{sys}) can be compressed by the Lipschitz constants $l_1$ and $l_2$.

\subsection{Conventional Robust MPC}
Due to the unknown disturbance and the constraints, we first consider the conventional robust MPC to deal with the system. Define $N$ as the prediction horizon. At each sampling instant $t_k$, the conventional MPC solves an OCP to obtain an optimal control sequence $\mathbf{u}^*(t_k)=\{u^*(t_k|t_k),u^*(t_k+1|t_k), \cdots, u^*(t_k+N-1|t_k)\}$ and only the first element would be applied to the real system. The cost function is formulated over the finite horizon as
\begin{eqnarray}
J(\bar{x}(t_k),\mathbf{\bar{u}}(t_k),t_k) \!\!&=&\!\! \sum^{N-1}_{i=0} L(\bar{x}(t_k+i|t_k),\bar{u}(t_k+i|t_k)) \nonumber \\
&&+ V_f(\bar{x}(t_k+N|t_k)),
\end{eqnarray}
where $L(\bar{x},\bar{u}) = \|\bar{x}\|^2_Q + \|\bar{u}\|^2_R$ is the stage cost function,  $V_f(\bar{x})=\|\bar{x}\|^2_P$ is the terminal penalty cost function and $\bar{x}$ and $\bar{u}$ represent feasible states and control inputs. In the cost function, $Q$ and $P$ are positive semi-definite matrices and $R$ is a positive definite matrix. Then, we construct the optimal control problem $\mathbf{1}$ as\\
\textbf{The OCP 1}:
\begin{eqnarray}
\mathbf{u}^*(t_k) = {\min_{\mathbf{\bar{u}}(t_k)\in \mathbb{U}}} J(\bar{x}(t_k),\mathbf{\bar{u}}(t_k),t_k),
\end{eqnarray}
subject to
\begin{subequations}
\begin{align}
&\bar{x}(t_k|t_k) = x(t_k),     \\
&\bar{x}(t_k+i+1|t_k) = f(\bar{x}(t_k+i|t_k),\bar{u}(t_k+i|t_k)),\\
&\bar{x}(t_k+i|t_k) \in \mathbb{X}\ominus \mathbb{X}_e(i), \quad \bar{u}(t_k+i|t_k)\in \mathbb{U}, \\
&\bar{x}(t_k+N|t_k) \in \mathbb{X}_\epsilon.
\end{align}
\end{subequations}
where $\mathbb{X}_e(i) = \{x:\|\bar{x}\|\leqslant i\eta(1+l)^i\}$ is the tightened state constraint set to guarantee the robustness of the system, $\mathbb{X}_\epsilon = \{x:\|\bar{x}\|_P\leqslant \epsilon, \epsilon>0\}$ is the robust terminal region and $\ominus$ means Minkowski subtraction of the sets.

For the nominal system (\ref{nonminalsys}), MPC is of recursive feasibility and closed-loop stability~\cite{chen1998quasi,luo2019robust} if: (i) the OCP has a feasible solution at the initial instant $t_0$; (ii) there is a local stabilizing controller $\kappa_f(x)$ in the robust terminal region to satisfy the constraint $\forall x \in \mathbb{X}_\epsilon, \kappa_f(x) \in \mathbb{U}$ such that:
\begin{eqnarray}
V_f(\bar{x}(t+1))-V_f(\bar{x}(t))\leqslant -L(\bar{x}(t),\kappa_f(\bar{x}(t))).
\end{eqnarray}
Moreover, if the controller is chosen as $\kappa_f(x) = Kx$ in linear systems, we have the Lyapunov equation of the weight matrices $Q$, $R$ and $P$, that is
\begin{eqnarray}\label{riccadi}
(A+BK)^TP+P(A+BK)\leqslant -Q^*,
\end{eqnarray}
where $Q^*=Q+K^TRK$, to solve the feedback gain $K$.

For conventional robust MPC, the heavy computation from solving OCP causes the mismatch of the sampling real states and the control input. Thus, how to ensure the one-to-one correspondence of the real system states and the optimal control input under the framework of MPC is the key improvement of this paper. In this paper, we intend to eliminate the delay impact by two ways: linearize the nonlinear system to reduce the computational complexity and compute the OCP ahead by prediction to change the pace, which is detailed in section \ref{sec:method}.

%% file: 04method.tex
\section{Method}
\label{sec:method}
In this section, we combine the piece-wise linearization and the prediction of the next system states to change the conventional MPC to deal with the delay problem. 

\subsection{Linearization of Nonlinear Systems}
Firstly, we linearize the nonlinear system model (\ref{sys}) to reduce computational complexity and fasten response speed with the MPC controller. Thus, at each control interval,  $t\in[t_k,t_{k+1}]$, by the second-order expansion of Taylor Polynomial, the system model $f(x(t),u(t))$ holds:
\begin{eqnarray}\label{linearation}
f(x(t),u(t)) \!\!\!\!\!\!&=&\!\!\!\!\!\! f(x(t_k),u(t_k)) + A_{t_k}\big[x(t)-x(t_k)\big] \nonumber \\
&+&\!\!\!\!\!\!B_{t_k}\big[u(t)-u(t_k)\big]  +R(x(t_k),u(t_k)) \nonumber \\
&=&\!\!\!\!\!\! A_{t_k}x(t)\!\!+\!\!B_{t_k}u(t) \!\!+\!\! \Omega \!\!+\!R(x(t_k),u(t_k)).
\end{eqnarray}
where $\Omega$ is the expansion error at the point $(x(t_k),u(t_k))$,
$\Omega = f(x(t_k),u(t_k))-(A_{t_k}x(t_k)+B_{t_k}u(t_k))$, and $R(x(t_k),u(t_k))$ is the Lagrange Remainder of the linearization error as:
\begin{eqnarray}
R(x(t_k),u(t_k)) = H_{x(t-k),u(t-k)}(x(t),u(t))\cdot \nonumber \\ f[x(t_k)+\theta_1(x(t)-x(t_k)),u(t_k)+\theta_2(u(t)-u(t_k))]
\end{eqnarray}
where $H$ is the Hessian matrix of the pair $(x(t_k),u(t_k))$, $\theta_1,\theta_2\in(0,1)$ are constants. Thus, by the Lagrange Mean Value Theorem, we can obtain the bound of the Hessian matrix at each piece interval, which means that the linearization error of the nonlinear system is bounded as
\begin{equation}
\label{linearbound}
\begin{aligned}
\|\Omega+R(x(t_k),u(t_k))\| \leqslant \|\Omega\|+ \quad\quad\quad \\
\eta_H\Big[l_1\|x(t)-x(t_k)\| +l_2\|u(t)-u(t_k)\|\Big] \triangleq \eta_2.
\end{aligned}
\end{equation}

Recalling the system (\ref{sys}), we can add this linearization error to the additional disturbances as a total disturbance. Thus, the total disturbances $w_t$ of the system model contains two parts with an upper bound $\eta$, that is
\begin{eqnarray}
\|w_t\| &=& \|e(t)\| + \|\Omega + R(x(t),u(t))\|, \nonumber\\
&\leqslant& \eta_1 + \eta_2 \triangleq \eta
\end{eqnarray}
which supplies the theoretical reliability for transforming the perturbed nonlinear system into a linear system with bigger bounded disturbances.

\subsection{Prediction of The Next Real States}
First, we redefine the control period to adapt the computational delay as
\begin{eqnarray}\label{delaydef}
t_{k+1}-t_k = \Delta t =m\delta, \quad m\in\mathbb{N}_{\geqslant 1},\quad m \leqslant T/\Delta t,
\end{eqnarray}
where $\delta$ is the minimal sampling interval. This equation (\ref{delaydef}) supposes that the computational time is an integral multiple of the sampling interval and less than the predictive horizon. By the last subsection, the system can be modelled in the interval $t\in[t_k,t_{k+1}]$ as
\begin{eqnarray}
x(t+1) = A_{t_k}x(t) + B_{t_k}u(t) + w_t.
\end{eqnarray}

Based on the current system states and the nominal system, we can predict the region of the m-steps future states $x(t_k+m)$ by the upper bound of the total disturbances and the nominal states $x^*(t_k+m|t_k)$. By the same control input sequence $\mathbf{\bar{u}}(t_k)$ and the recursion of system dynamics, the state deviation between the nominal system and the real system from $t_k$ to $t_k+m$ is bounded by
\begin{eqnarray}
\|x_e(t_k+m)\|\leqslant \frac{\bar{\lambda}(A)^m-1}{\bar{\lambda}(A)-1}\eta, \quad m \in [0,T],
\end{eqnarray}
where $\bar{\lambda}(A)$ is the maximum eigenvalue of linear system matrix $A$, $\eta$ is the total disturbances and $x_e(t_k+m) \triangleq x(t_k+m)-x^*(t_k+m|t_k)$ is the deviation between the nominal system and the real disturbed system. As the upper bound of the accumulative state deviation is an $e$-exponential form of the control interval time $m\delta$, we can predict the neighbour region of the next real triggering state $x(t_{k+1})$ by the current real states $x(t_k)$ and the bound of disturbances $\eta$, which is the key to compute the OCP ahead. By triangle inequality, the real system states can be bounded as
\begin{equation}\label{realtran}
\|x(t_k+m)\|\leqslant\|x^*(t_k+m|t_k)\|+\frac{\bar{\lambda}(A)^m-1}{\bar{\lambda}(A)-1}\eta.
\end{equation}
Referring to the definition of a disturbed invariant set, we can replace (\ref{realtran}) as
\begin{equation}\label{disturbanceinvar}
\|x(t_k+m)\|\in\mathbb{X}_\omega(x^*(t_k),m),
\end{equation}
where $\mathbb{X}_\omega(x^*(t_k),m)$ means an ellipsoid with the center $x^*(t_k+m|t_k)$ and the radius $\frac{\bar{\lambda}(A)^m-1}{\bar{\lambda}(A)-1}\eta$.

\subsection{Robust Tube-based Smooth MPC}
After the linearization of the nonlinear system and the prediction of the region of the next real system state, our controller can compute the OCP ahead. In this subsection, the controller $u(t_k)$ is designed as
\begin{eqnarray}\label{tubempc}
u(t|t_k) = v(t|t_k) + K[x(t)-x^*(t|t_k)],
\end{eqnarray}
where $v(t_k)$ is the nominal optimal control decision variable for the linear system and $K$ is the state feed-back gain computed by the Riccati Equation as (\ref{riccadi}) to restrain the bounded disturbances. Referring to the conventional MPC, we set the optimal control problem $\mathbf{2}$, which is formulated as\\
\textbf{The OCP 2}:
\begin{eqnarray}
\mathbf{v}^*(t|t_k) = {\min_{\mathbf{\bar{u}}(t|t_k)\in \mathbb{U}}} J(\bar{x}(t|t_k),\mathbf{\bar{v}}(t|t_k),t_k),
\end{eqnarray}
subject to
\begin{subequations}
\begin{align}
&\bar{x}(t_{k+1}|t_k) \in \mathbb{X}_{\omega}(z^*(t_k),m),     \\
&\dot{\bar{x}}(t|t_k) = A_{t_k}\bar{x}(t|t_k)+B_{t_k}\bar{u}(t|t_k),\\
&\bar{x}(t|t_k) \in \mathbb{X}\ominus \mathbb{X}_e(t),\ \ \bar{u}(t|t_k)\in \mathbb{U}\ominus K\bar{x}, \\
&\bar{x}(t_k+T|t_k) \in \mathbb{X}_\epsilon,\quad t\in[t_{k+1},t_{k+1}+T]. \label{terminal}
\end{align}
\end{subequations}
where $J(\bar{x}(t|t_k),\mathbf{\bar{v}}(t|t_k),t_k)$, $\mathbb{X}_e(i)$ and $\mathbb{X}_\epsilon$ have the same definition with the OCP 1. By solving the OCP 2, we can obtain the optimal control sequence $\mathbf{v}^*(t_k)$. Then, the first $m$ elements are used to predict the optimal state $x^*(t_k+m|t_k)$ and stored until the next real system states come. At the next computational instant, we can apply $u^*(t_k)$ to the system and repeat this process until the system converges. Based on the theoretical framework of tube MPC, we implement our control strategy in Algorithm~\ref{def_alg1}.


\begin{algorithm}[htpb]
\caption{Robust tube-based smooth MPC}
\label{def_alg1}
\begin{algorithmic}[1]
\State \emph{Offline}: Initialize the parameters $m$, $l$ of system (\ref{sys}) and set the weight matrices $Q$ and $R$. By (\ref{riccadi}), compute the terminal state feed-back gain $K$ and the weight matrix $P$. Then, define the terminal set to satisfy (\ref{terminal}). Find the optimal solution $v^*(t_0)$ for the initial state $x(t_0)$.
\State \emph{Online}:\\ for each triggering time $t_k, k=1,2,3,\ldots$
\State \quad \quad (i) Apply the first $m$ elements of the optimal control sequence $v^*(t_{k-1})$ to (\ref{tubempc}) for the real system.
\State \quad \quad (ii) Measure the current state $x(t_k)$ and compute the predictive state $x^*(t_k+m|t_k)$ by the nominal nonlinear model (\ref{nonminalsys}) and the last optimal control sequence $v^*(t_{k-1})$. Then, estimate the  predictive disturbed state set $\mathbb{X}_{\omega}(x^*(t_k),m)$.
\State \quad \quad (iii) Based on the state $x^*(m|t_k)$, linearize the model (\ref{nonminalsys}) to obtain system matrices $A_{t_k}$ and $B_{t_k}$ and compute the local state feed-back gain $K_{t_k}$ as (\ref{riccadi}).
\State \quad \quad (iv) Solve the OCP 2 to obtain the optimal control sequence $v^*(t_k)$ for the next triggering instant $t_{k+1}$.
\State \quad \quad ((ii) to (iv) are synchronous with (i))
\State \quad \quad (v) Let $k = k+1$.
\end{algorithmic}
\end{algorithm}


%% file: 05experiment.tex
\section{Experiments}
\label{sec:exp}
In this section, we present a comprehensive series of experiments, combining numerical simulations and practical implementation on a physical robot platform. The objective of these experiments is to provide empirical evidence of the effectiveness of our proposed algorithm, showcasing its superiority in terms of control performance and responsiveness compared to established MPC controllers.

Our evaluation includes a comparative analysis of the following control strategies:
(1)\textbf{Ideal-Continuous MPC}: This controller represents the theoretical best-case scenario exhibiting optimal control performance with no computation delay at each sampling instant, (2)\textbf{Time-Triggered MPC}: As a commonly employed control strategy in real systems, this controller introduces the worst-case delay for each control period, reflecting real-world scenarios, (3)\textbf{Proposed Robust Tube-Based Smooth-MPC}: Our novel approach utilizes piece-wise linearization and state prediction for enhanced control performance and smoothness.

Numerical simulation experiments are conducted using a UR5 robot arm in, Rviz simulator, focusing on two typical manipulation tasks: reaching a specified point and tracking a composite trajectory. Moreover, we extend our evaluation to a physical robot manipulator, implementing the same tasks in a real-world setting. This practical experimentation further reinforces the algorithm's viability and applicability to real-world robotic systems. The combination of numerical simulations and physical experiments provides a comprehensive validation of the proposed algorithm's capabilities.
\subsection{Numerical Simulations}
\begin{figure}[t]
	\centering
	\includegraphics[width=1.0\linewidth]{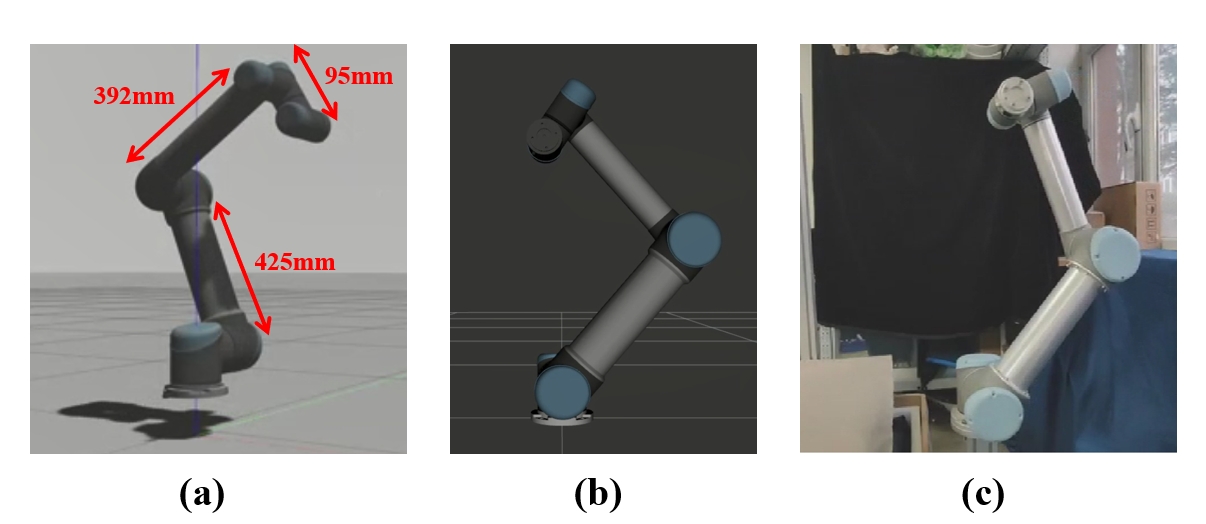}
	\caption{(a) The UR5 manipulator in the Rviz simulator along with its geometries. (b) The manipulator moves within the plane. (c) The lateral view of the UR5 manipulator in the real environment.}
	\label{fig:simulation}
 \vspace{-10pt}
\end{figure}

To simplify computations, we lock the 1st, 5th, and 6th joints of the robotic arm. Thus, we can build our model with a three-joint robot manipulator operating within the $x-y$ plane shown in Fig. \ref{fig:simulation} (a) and (b). The manipulator features three revolute joints and employs three angular velocity controls for plane motion. 
The kinematic dynamics of the manipulator's end-effector point are described by the following system equations:
\begin{eqnarray}\label{realsys}
  \begin{bmatrix} \dot{x} \\ \dot{y} \\ \dot{\theta_1} \\ \dot{\theta_2} \\ \dot{\theta_3} \end{bmatrix}=
  \begin{bmatrix} -l_1 s\theta_1 & -l_2 s\theta_2 & -l_3 s\theta_3 \\ l_1 c\theta_1 & l_2 c\theta_2 & l_3 c\theta_3 \\
  1 & 0 & 0 \\ 0 & 1 & 0 \\ 0 & 0 & 1
  \end{bmatrix}
  \begin{bmatrix} \omega_1 \\ \omega_2 \\ \omega_3 \end{bmatrix} + \begin{bmatrix} e_1 \\ e_2 \\ e_3 \\ e_4\\ e_5 \end{bmatrix}
\end{eqnarray}
Here, $s,c$ denote trigonometric functions $\sin(\cdot),\cos(\cdot)$. The system states contain $(x(t), y(t), \theta_1(t), \theta_2(t), \theta_3(t))$ and the control inputs are $(\omega_1(t),\omega_2(t),\omega_3(t))$. The coordinates of the end-effector are denoted by $p(t) = (x(t),y(t))$, and $(\theta_1(t),\theta_2(t),\theta_3(t))$,$(\omega_1(t),\omega_2(t),\omega_3(t))$ signifies the joint angles and angular velocities. The lengths of the manipulator's three links are represented by $l_1$, $l_2$, $l_3$.

In the numeric simulation, we set $L_1 = L_2 = \sqrt{5}, L_3 = \sqrt{10}$ and the system constraints are defined as $\frac{\pi}{2}\leqslant\theta_1\leqslant\pi, 0\leqslant\theta_2\leqslant\pi, 0\leqslant\theta_3\leqslant\frac{\pi}{2}$ for the state and $-\frac{\pi}{16}\leqslant\omega_1,\omega_2,\omega_3\leqslant\frac{\pi}{16}$ for the control input. For all MPC controllers in our experiments, we employ a common sampling time $\delta = 0.1$ along with a prediction horizon of $T=3$. To shape the controller's behaviour, we define the weight matrices as $Q = 0.1 \textbf{\emph{I}}^{5\times5}$ and $R = 0.01 \textbf{\emph{I}}^{3\times3}$, where $\textbf{\emph{I}}$ represents the identity matrix. Based on numerous trial runs, we ascertain that the computation time for solving the nonlinear OCP 1 is consistently recorded as $m=2.8s$, while the computation time for solving the linear OCP 2 is $\Delta t=0.3s$. The simulation examples are implemented using YALMIP toolbox on Matlab 2020a with intel$^\circledR$ core$^{\text{TM}}$ i9-9900 CPU.
\begin{figure}[t]
	\centering
	\includegraphics[width=1.0\linewidth]{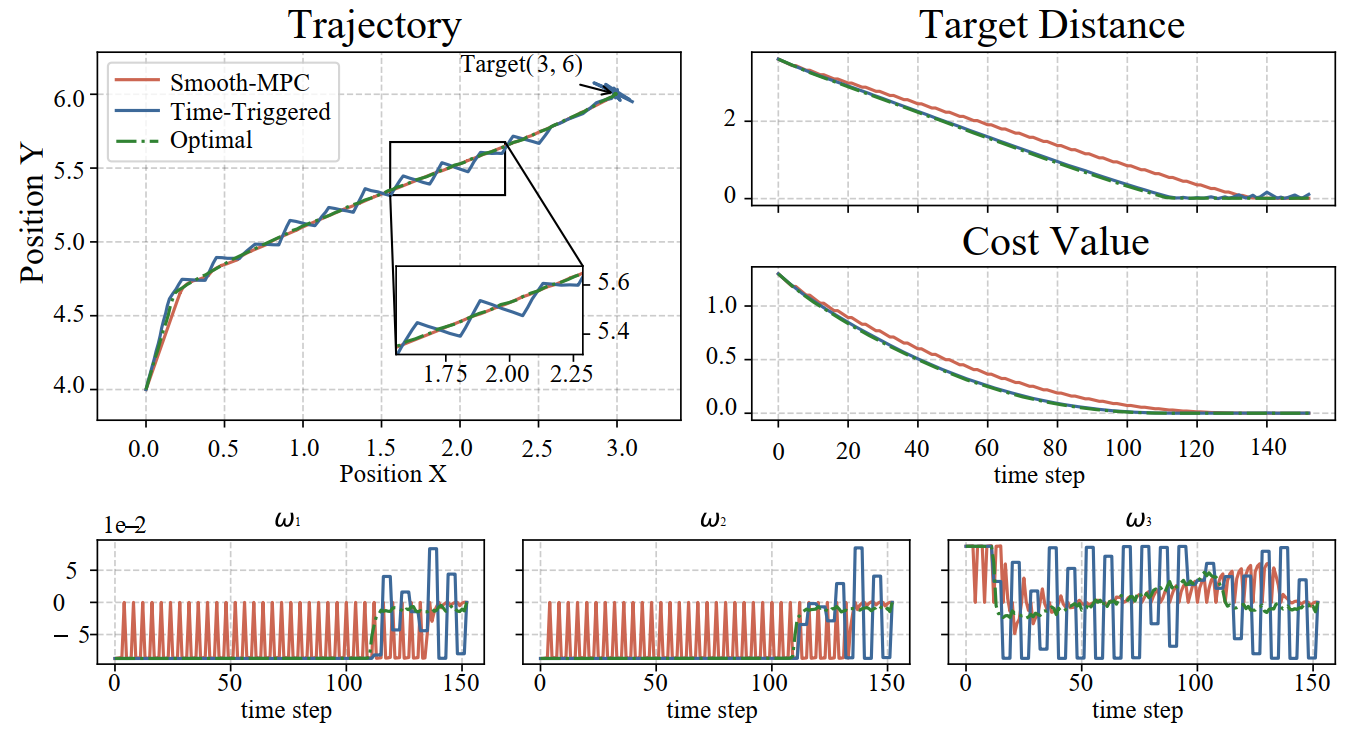}
	\caption{The comparison of three control methods over the Position Tracking task, which contain the state trajectory, position error, cost function value and control input.}
	\label{Position_Tracking}
  \vspace{-10pt}
\end{figure}
\begin{figure}[t]
	\centering
	\includegraphics[width=1.0\linewidth]{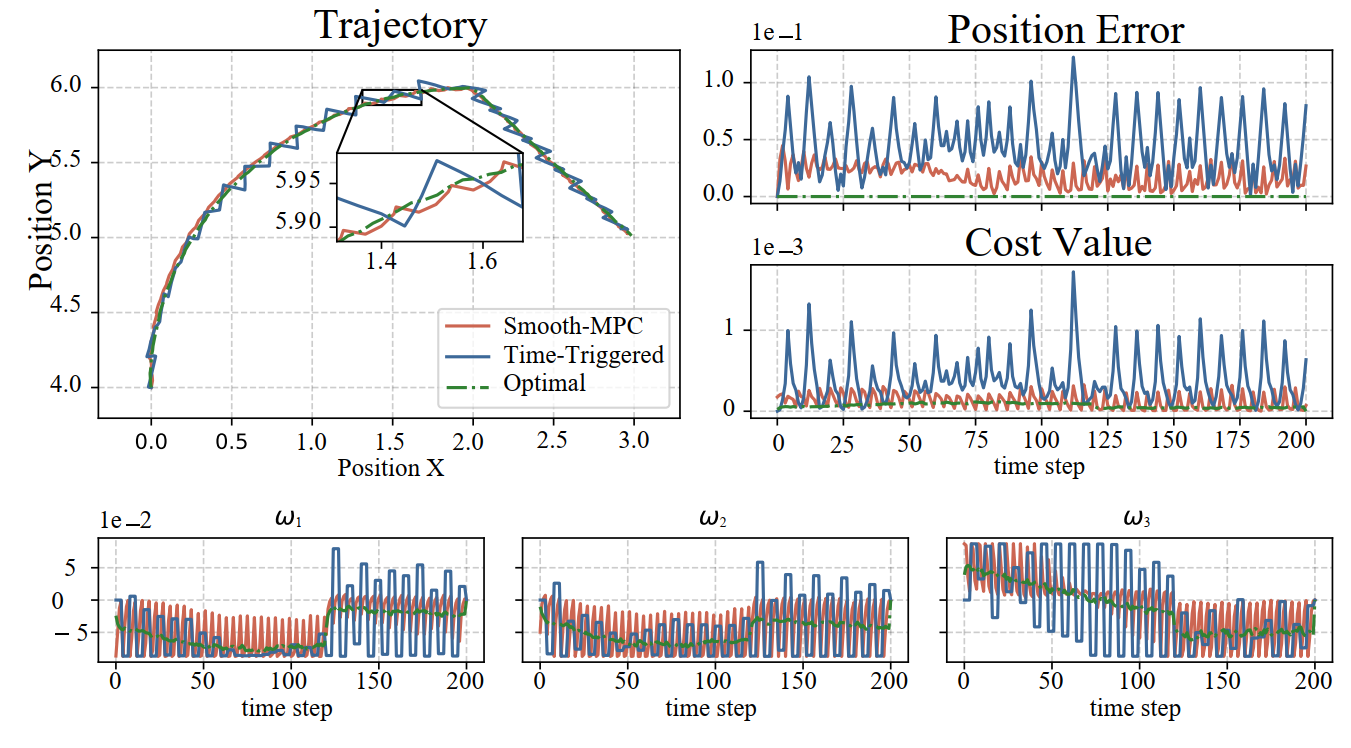}
	\caption{The comparison of three control methods over the Trajectory Tracking task, which contain the state trajectory, position error, cost function value and control input.}
	\label{Trajectory_Tracking}
 \vspace{-10pt}
\end{figure}

\textbf{Position Reaching:} In this task, we evaluate the performance of the proposed algorithms in a position-reaching task, where the manipulator is tasked with smoothly moving an object from the initial position $(0,4)$ to the target position $(2,6)$. We compare the outcomes of the optimal, time-triggered, and smooth MPC controllers for this task.

As depicted in Figure \ref{Position_Tracking}, the trajectories of the endpoints are shown for each controller. The figures reveal that our proposed methods exhibit superior reaching performance when compared to the time-triggered MPC. Notably, both the optimal and our proposed methods produce smoother trajectories, showcasing improved control precision and effectiveness.

\textbf{Trajectory Tracking:} In this task, the endpoint tracks a specific trajectory with multiple splices, consisting of a quarter circle and an oblique line. We use this task to emulate complex manipulation processes with given spline curves. From Fig.\ref{Trajectory_Tracking}, our approach shows better tracking performance, higher robustness and significantly lower position errors than the time-triggered one. Besides, our control inputs have a relatively low amplitude, which can mitigate the effects of input jumps.
\subsection{Physical Experiments}
Similar to numerical simulations, we compare our proposed control strategy with two baseline methods in a real-world setting with the same type UR5 manipulator shown in Fig \ref{fig:simulation} (c). In every test episode, the manipulator conducts a series of steps to achieve the target position with a joint angel velocity combination computed by different MPC methods. The optimized trajectories are presented as experimental results to compare the performance of different methods. During the validation, we record the signal of the unlocked 2nd, 3rd, and 4th joints of our manipulator platform and transform the collected joint data into the trajectory of the end-effector.\par
\begin{figure}[t]
    \centering
    \includegraphics[width=1.0\linewidth]{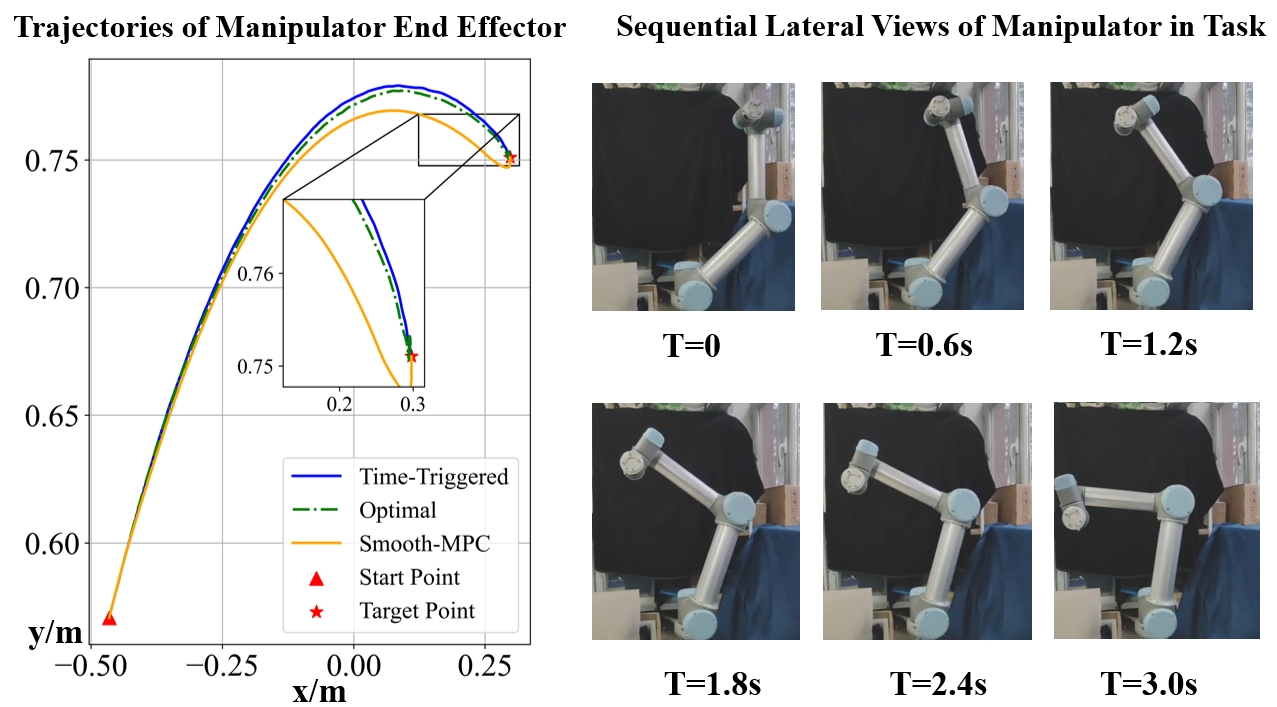}
    \caption{The trajectories of the end-effector in position tracking task with different control strategies and video frames of experimental scenario.}
    \label{fig:real-position}
    \vspace{-10pt}
\end{figure}

The results of the task moving to the target position are shown in  Fig. \ref{fig:real-position}. Our proposed method outperforms the baselines in both aspects of convergence and robustness.
It is worth noting that, in the above experimental results, the ``Optimal'' method is merely the theoretically calculated optimal solution. In the physical experiments presented in this paper, the ``Optimal'' method is implemented by dynamically computing the next move at each moment based on the current motion state. 
\begin{table}[t]
\caption{A summary of comparison for computation time. \textbf{Single solution}: the time of a single optimal control problem; \textbf{Whole solution}: the time of whole control process.}
\label{computation_time}
\resizebox{\columnwidth}{!}{
\begin{tabular}{llll}
\hline
& Single solution & Whole solution & Percentage  \\ \hline
Time-triggered MPC & 2.8s & 840s & $100\%$ \\
Optimal MPC w/o delay & 2.5s & 750s & $89\%$ \\
\textbf{Ours} & 0.3s & 90s & $12\%$ \\ \hline
\end{tabular}
}
\end{table}

Besides, we also conduct the trajectory following task in a physical setting, the results are shown in Fig. \ref{fig:real-trajectory}. In the trajectory tracking task, our proposed Smooth MPC method demonstrates the capability to accurately follow the predefined trajectory with relatively reduced oscillations. To illustrate the real-time performance of our algorithm, We report the computation time under three methods, as shown in Table~\ref{computation_time}, containing the time comparison within a single solution for an optimal control problem and the whole control process.  The results verify that our proposed method demonstrates efficient computational performance while ensuring manipulation accuracy and control performance.